\documentclass[sigconf]{aamas}

\usepackage{balance} 
\usepackage{algorithm}
\usepackage{algorithmic}

\usepackage{fontawesome}
\usepackage{booktabs}
\usepackage{array}
\usepackage{xcolor}
\usepackage{colortbl}
\usepackage{multirow}

\definecolor{headerblue}{RGB}{52, 73, 94}
\definecolor{lightgray}{RGB}{248, 249, 250}
\definecolor{mediumgray}{RGB}{233, 236, 239}

\doi{EKAW8904}

\makeatletter
\gdef\@copyrightpermission{
  \begin{minipage}{0.2\columnwidth}
   \href{https://creativecommons.org/licenses/by/4.0/}{\includegraphics[width=0.90\textwidth]{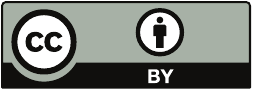}}
  \end{minipage}\hfill
  \begin{minipage}{0.8\columnwidth}
   \href{https://creativecommons.org/licenses/by/4.0/}{This work is licensed under a Creative Commons Attribution International 4.0 License.}
  \end{minipage}
  \vspace{5pt}
}
\makeatother

\setcopyright{ifaamas}
\acmConference[AAMAS '26]{Proc.\@ of the 25th International Conference
on Autonomous Agents and Multiagent Systems (AAMAS 2026)}{May 25 -- 29, 2026}
{Paphos, Cyprus}{C.~Amato, L.~Dennis, V.~Mascardi, J.~Thangarajah (eds.)}
\copyrightyear{2026}
\acmYear{2026}
\acmDOI{}
\acmPrice{}
\acmISBN{}

\title[LLMArg]{Argumentative Human-AI Decision-Making: \\ Toward AI Agents That Reason With Us, Not For Us}
\subtitle{Blue Sky Ideas Track}

\author{Stylianos Loukas Vasileiou}
\affiliation{
  \institution{New Mexico State University}
  \city{Las Cruces, NM}
  \country{USA}
  }
\email{stelios@nmsu.edu}

\author{Antonio Rago}
\affiliation{
  \institution{King's College London}
  \city{London}
  \country{UK}
  }
\email{antonio.rago@kcl.ac.uk}

\author{Francesca Toni}
\affiliation{
  \institution{Imperial College London}
  \city{London}
  \country{UK}
  }
\email{f.toni@imperial.ac.uk}

\author{William Yeoh}
\affiliation{
  \institution{Washington University in St. Louis}
  \city{St. Louis, MO}
  \country{USA}
  }
\email{wyeoh@wustl.edu}

\begin{abstract}
Computational argumentation offers formal frameworks for transparent, verifiable reasoning but has traditionally been limited by its reliance on domain-specific information and extensive feature engineering. In contrast, LLMs excel at processing unstructured text, yet their opaque nature makes their reasoning difficult to evaluate and trust. We argue that the convergence of these fields will lay the foundation for a new paradigm: \textit{Argumentative Human-AI Decision-Making}. We analyze how the synergy of \textit{argumentation framework mining}, \textit{argumentation framework synthesis}, and \textit{argumentative reasoning} enables agents that do not just justify decisions, but engage in dialectical processes where decisions are contestable and revisable -- reasoning \textit{with} humans rather than \textit{for} them. This convergence of computational argumentation and LLMs is essential for human-aware, trustworthy AI in high-stakes domains.
\end{abstract}


\newcommand{\BibTeX}{\rm B\kern-.05em{\sc i\kern-.025em b}\kern-.08em\TeX}

\begin{document}


\pagestyle{fancy}
\fancyhead{}


\maketitle 
\allowdisplaybreaks
\sloppy


\section{Introduction}

For over three decades, computational argumentation (CA) has offered formal frameworks for valid and sound reasoning, enabling transparent and verifiable decision-making through approaches such as abstract \cite{dung1995,Baroni_18} and structured \cite{besnard2014constructing} argumentation. However, its practical application has often been constrained by the need for structured inputs, such as hand-crafted knowledge bases and domain-specific rules. On the other hand, large language models (LLMs) have demonstrated an unprecedented ability to process and generate natural language \cite{NEURIPS2020_1457c0d6}. Yet, their internal reasoning remains approximate and largely opaque, making their outputs difficult to verify and trust \cite{mccoy2023embers,valmeekam2023planning,beyer-reed-2025-lexical,gemechu2025natural}.

The integration of CA and LLMs presents an opportunity to address their complementary limitations.
LLMs gain structured reasoning through formal argumentation frameworks (e.g., graphs of interconnected arguments with well-defined semantics), while CA overcomes its reliance on extensive knowledge engineering and gains the ability to operate at scale on unstructured text \cite{Bezou-Vrakatseli_24}. Recent systems already demonstrate the power of this integration by using LLMs to construct formal argumentation frameworks for tasks such as explainable and contestable claim verification, which are then evaluated by a formal argumentation engine \cite{freedman2025argumentative,zhu2025argrag}.

We propose that the convergence of CA and LLMs enables a new paradigm: AI agents that engage in dialectical reasoning processes with humans. We organize this vision by first presenting a taxonomy of three core tasks: \textit{argumentation mining}, \textit{argumentation synthesis}, and \textit{argumentative reasoning}. 
We then examine recent developments in each area, demonstrating how the integration of LLMs with CA is not merely enhancing existing capabilities but creating fundamentally new approaches to human-AI collaboration. These synergistic developments signal the emergence of interactive, contestable AI agents, giving systems that reason with humans rather than for them.
%
Such agents 
will mark the emergence of \textit{argumentative human-AI decision-making}.

\section{Taxonomy of Argumentation Tasks}
\label{sec:taxonomy}

Traditionally, CA has centered on three core tasks related to argumentation frameworks (AFs): \textit{argumentation mining} (extracting formal argument structures from natural text), \textit{argumentation synthesis} (generating new arguments within formal constraints), and \textit{argumentative reasoning} (evaluating and explaining arguments through formal argumentation). 
%
%
In the remainder of this section, we analyze developments across these tasks, summarized in Table~\ref{tab:taxonomy}.

\subsection{Task I: Argumentation Mining}

Argumentation mining is the task of automatically identifying and extracting AFs from unstructured text \cite{lawrence2020argument}. The application of LLMs to this task helps bridge the gap between raw natural language and the structured representations needed for formal reasoning. This eliminates the need for many of the handcrafted features and domain-specific heuristics that limited traditional approaches.

\noindent \textbf{AF Extraction} AF extraction identifies argumentative units (e.g.,~premises, claims) and the relations (e.g., support, attack) between them. 
For instance, given ``The proposed method outperforms the baseline because it reduces inference time by 40\%,'' extraction identifies ``reduces inference time by 40\%'' as a premise supporting the claim ``the proposed method outperforms the baseline.''
While traditional transformer baselines like RoBERTa established strong performance \cite{ruiz2021transformer}, they necessitated task-specific training. In contrast, LLMs have shifted the paradigm toward \emph{in-context learning}, allowing general-purpose models to capture complex argumentative relations via few-shot prompting alone \cite{gorur2025large}. Beyond prompting, recent work frames extraction as a text generation task: \citet{cabessa2025argument} achieved state-of-the-art results by fine-tuning open-weight models, while the ArgInstruct framework \cite{Stahl2025ArgInstructSI} utilizes instruction tuning to handle unseen tasks in zero-shot settings, effectively unifying fragmented extraction pipelines.

\begin{table}[t!]
\centering
\small
\renewcommand{\arraystretch}{1.5}
\setlength{\tabcolsep}{2pt}
\resizebox{\columnwidth}{!}{
\begin{tabular}{@{}p{2.0cm}p{6.0cm}p{1.5cm}@{}}
\rowcolor{headerblue}
\textcolor{white}{\textbf{Task}} &
\textcolor{white}{\textbf{Description}} &
\textcolor{white}{\textbf{Representa-tive Work}} \\ 
\midrule
\rowcolor{gray!8}
\textbf{Argumentation Mining} & 
\textbf{Definition:} Automatic identification and extraction of argumentative structures from text \newline
\textbf{Enhancement:} \textit{LLMs} $\rightarrow$ Eliminating need for manual features and structured inputs \newline
\textbf{Subtasks}: \textit{AF Extraction}, \textit{Content Detection} &
\cite{cabessa2025argument, ruiz2021transformer, gorur2025large, Stahl2025ArgInstructSI, rocha2024cross, lan2024stance, li2025large, bezou2025can} \\
\midrule
\textbf{Argumentation Synthesis} & 
\textbf{Definition:} Generation of new arguments, premises, claims, and summaries \newline
\textbf{Enhancement:} \textit{LLMs} $\rightarrow$ Enabling flexible, scalable generation without templates \newline
\textbf{Subtasks}: \textit{AF Generation}, \textit{AF Summarization} &
\cite{chen2024exploring, gray2025generating, mouchel-etal-2025-logical, hu-etal-2024-americano-argument, altemeyer2025argument, roush2024opendebateevidence, Li2024WhichSAA} \\
\midrule
\rowcolor{gray!8}
\textbf{Argumentative Reasoning} & 
\textbf{Definition:} Application of computational argumentation for sound decision-making \newline
\textbf{Enhancement:} \textit{Argumentation} $\rightarrow$ Providing formal semantics and transparent, contestable reasoning \newline
\textbf{Subtasks}:~\textit{Claim Verification}, \textit{Explainable Decision-Making} &
\cite{yue2024retrieval, si_checkwhy, freedman2025argumentative, ng2025marge, zhu2025argrag, hong2024argmed, lin2024towards} \\
\bottomrule
\end{tabular}
}
\caption{A taxonomy of synergistic tasks for LLMs and CA.} 
\label{tab:taxonomy}
\end{table}

\noindent \textbf{Content Detection} Content detection focuses on identifying specific properties of argumentative text such as the author's stance (e.g.,~supporting/opposing a position), identifying reasoning patterns (analogical, causal), or detecting linguistic markers. For the same example above, content detection would identify the author's positive stance toward the method. Traditional pipelines typically trained single-task classifiers, often with limited cross-genre transfer. LLMs address this limitation through agentic and prompt-based formulations. \citet{lan2024stance} introduced COLA, which reframes stance detection not as static classification, but as a collaborative multi-agent debate, achieving state-of-the-art performance without labeled data. For cross-genre robustness, \citet{rocha2024cross} utilized LLMs to insert missing discourse markers, making implicit relations explicit. Furthermore, the larger LLMs have demonstrated strong few-shot capabilities in classifying complex argument schemes, a task where smaller LLMs remain inadequate \cite{bezou2025can}.


\subsection{Task II: Argumentation Synthesis}

Argumentation synthesis involves generating new AFs, such as premises, claims, counterarguments, and summaries \cite{Reed_generating_arguments,carenini2006generating,green2011natural}. 
Earlier systems relied on templates or rigid generation pipelines that limited flexibility \cite{sato2015end,hua2019argument,el2019computational}. Like mining, synthesis is enhanced by the generative power of LLMs, which allows for the flexible creation of new AFs and summaries from natural language prompts.

\noindent \textbf{AF Generation} AF generation concerns the creation of new argumentative units that are grammatically correct and logically sound. This requires three core capabilities: understanding the topic, organizing premises and claims into coherent argumentative relations, and ensuring adherence to well-defined argumentation schemes. Historically, generating arguments relied on templates and domain-specific pipelines, often resulting in repetitive or unconvincing outputs. LLMs surmount these limitations by enabling multi-step reasoning and instruction adherence. \citet{chen2024exploring} validated this potential, showing strong performance across generation tasks. Recent architectures further improve quality through agentic interaction: AMERICANO \citep{hu-etal-2024-americano-argument} employs a generate-critique-refine loop to ensure factual grounding. Furthermore, researchers are now addressing the logical reliability of these generations. \citet{mouchel-etal-2025-logical} introduced preference-based fine-tuning to penalize fallacious reasoning, and \citet{gray2025generating} utilized scheme-based prompting to enforce valid legal argumentation patterns.

\noindent \textbf{AF Summarization} Summarizing argumentative discourse requires preserving central claims and key points of contention. Earlier methods struggled with coverage and salience in long or heterogeneous corpora. With LLMs and large-scale resources such as OpenDebateEvidence \citep{roush2024opendebateevidence}, recent systems achieve better results. \citet{altemeyer2025argument} integrated LLMs into AF summarization and evaluation, reporting substantial gains over traditional approaches. However, end-to-end pipelines remain challenging. \citet{Li2024WhichSAA} introduced a multi-task dataset covering the full workflow of preparing an argumentative essay and evaluated multiple generative baselines. Their findings reveal that while LLMs perform well on individual subtasks, performance degrades when tasks are chained together, highlighting the problem of error propagation in complex pipelines.

\subsection{Task III: Argumentative Reasoning}

Argumentative reasoning applies CA semantics for decision-making, verification, and explanation generation \cite{amgoud2009argumentation,atkinson2017towards,Cyras_21}. While argumentation mining and synthesis are primarily enhanced by LLMs, argumentative reasoning inverts this relationship. In this task, LLMs handle the front-end challenge of mapping unstructured natural language onto formal AFs, while CA provides the back-end engine for transparent, robust, and verifiable reasoning. For example, while the LLM extracts the paper's claims, the formal reasoning engine determines if the paper's conclusion logically follows from the provided experimental results, or flags a contradiction between the methodology and the results. This hybrid approach overcomes the brittleness of traditional symbolic systems by enabling dynamic, explainable decision-making directly from textual data.

\noindent \textbf{Claim Verification} Claim verification assesses the truthfulness of a claim (or argument) by weighing conflicting evidence. While early rule-based approaches lacked open-domain adaptability \cite{Bindris2020ClaimCCA,Ahmadi2019ExplainableFCA,Estes2022FactCMA}, LLM-driven pipelines now enable verification through explicit argument construction. RAFTS \citep{yue2024retrieval} replaces opaque scoring with the synthesis of contrastive arguments, improving auditability, while the CHECKWHY benchmark \citep{si_checkwhy} enables multi-step (causal) reasoning for verdict justification. Importantly, recent work integrates formal semantics to enforce logical consistency. Systems such as Argumentative LLMs \citep{freedman2025argumentative} and ArgRAG \citep{zhu2025argrag} use LLMs to map claims/evidence into quantitative AFs, deferring the final decision to a deterministic solver. Consequently, MArgE \citep{ng2025marge} extended this neuro-symbolic approach to multi-model settings, meshing arguments from diverse LLMs to mitigate individual hallucinations.

\noindent \textbf{Explainable~Decision-Making} Explainability requires that decisions can be inspected and contested, with traditional approaches using handcrafted AFs (c.f. \cite{Cyras_21,Vassiliades_21} for overviews), as well as other forms of argumentative explanation resulting from them (c.f. \cite{Rago_24}). 
LLM-based approaches now generate natural language arguments and formalize them into AFs for transparent inference. 
Frameworks like Argumentative LLMs \citep{freedman2025argumentative} and ArgRAG \citep{zhu2025argrag} externalize reasoning into editable graphs, allowing users to explicitly add, strengthen, weaken, or remove arguments to observe outcome changes. This capability extends to specialized domains: ArgMed-Agents \citep{hong2024argmed} structures clinical discussions via argumentation schemes, while recent multimodal systems utilize multi-agent debate to provide explicit justifications for image-text classifications \citep{lin2024towards}.


Across these tasks, we observe the synergistic relationship between LLMs and CA. 
This synergy is beginning to enable capabilities neither approach could achieve independently, particularly the ability to construct, evaluate, and revise AFs through natural language interaction while maintaining formal guarantees. This is pointing toward our central thesis: the emergence of \textit{argumentative human-AI decision-making}.

\section{Toward Argumentative Human-AI Decision-Making}

The advances we have traced across the tasks in the previous section are converging toward AI agents that do not simply assist with decisions but engage with humans as argumentative partners. 
The full realization of argumentative human-AI collaboration will require agents that can propose and evaluate claims, surface and weigh evidence, and adapt their reasoning as humans add context or express preferences. The goal will not be to replace the humans in the loop, but to amplify them by making complex reasoning more transparent, contestable, and adaptable to domain norms. In what follows, we sketch what such AI agents might look like, and how they may reshape practice in some high‑stakes domains.

The transition from current systems to this vision requires bridging three gaps: moving from single-shot argument generation to iterative refinement, from explaining individual decisions to exposing entire reasoning processes, and from domain-agnostic models to systems aligned with professional norms. Current prototypes like Argumentative LLMs \cite{freedman2025argumentative} and ArgRAG \cite{zhu2025argrag} demonstrate feasibility but operate in constrained settings. The next generation of agents must handle open-domain reasoning while maintaining formal guarantees.

\noindent \textbf{Contestable Architectures} The defining property of these agents is contestability, that is, decisions must be open to inspection and revision. 
The architecture operates through three components. First, separation of generation and evaluation: the LLM generates candidate arguments and relations as structured outputs, while a formal argumentation solver determines acceptability. This ensures probabilistic content generation with deterministic inference. Second, end-to-end provenance, where each argument node contains pointers to source text spans, enabling auditing of hallucinatory or unsupported claims. Third, bidirectional propagation; when users modify the framework, e.g., adjusting argument strength, adding attack relations, or introducing new premises, the solver recomputes acceptability labels across the entire graph. For example, in medical diagnosis, an initial framework might conclude ``prescribe medication X'' based on arguments $A_1$ (symptoms match condition Y) and $A_2$ (X treats Y effectively). If a physician adds argument $A_3$ (patient has contraindication Z) attacking $A_2$, the solver propagates this change, potentially flipping the conclusion to ``avoid X, consider alternative.'' This is not only explanation, but collaborative reasoning through formal structures.

Note that our paradigm differs from traditional XAI. Current XAI methods provide post-hoc rationalizations, i.e., explaining what an opaque model did \cite{guidotti2018survey,belle2021principles}, or explaining why certain decisions were reached~\cite{vasileiou2021exploiting,SreedharanCK21,vasileiou2022jair,vasileiou2025generating}. Even interactive XAI systems that allow probing with ``what-if'' scenarios still treat the model's reasoning as a fixed black box. In contrast, our paradigm, which is similar in spirit to the Evaluative AI framework \cite{miller2023explainable},\footnote{The Evaluative AI framework aims to improve human decision-making by providing users with evidence for (or against) a decision, instead of a single recommendation.} externalizes reasoning into formal AFs where inference becomes the interface. In other words, XAI explains the product (a decision), while our approach makes the process (the reasoning) the primary artifact. 

\noindent \textbf{Interactive Revisions} While contestable architectures make reasoning inspectable, interactive revisions make it improvable. We envision agents that engage in iterative cycles of proposal, critique, and revision, as in \cite{Rago_23}. This is not just ``chatting'' with a model; it is a structured dialectical protocol where humans provide their reasons and normative constraints \cite{vasileiou2024dialectical}. The agent then regenerates the AF to satisfy these constraints while maintaining logical consistency. This enables role-aware collaboration: the human sets the strategy and value axioms, while the agent handles the combinatorial complexity of connecting claims to evidence. These interactions serve a dual purpose: they solve the immediate problem and provide feedback data to align the agent with domain-specific (reasoning) norms, such as legal standards of proof or clinical risk thresholds.

\noindent \textbf{Possible Applications} To understand why this paradigm shift matters, consider the following high-stakes applications in medicine and scientific peer review. 


In medicine, decisions require integrating heterogeneous evidence, such as clinical trials, patient histories, treatment guidelines, under significant uncertainty. Current clinical decision support systems often function as black boxes, providing recommendations without accessible reasoning. ArgMed-Agents \cite{hong2024argmed} point toward an alternative, namely AI agents that construct explicit argumentative justifications for clinical recommendations. When combined with contestable architectures, these become negotiable recommendations. An agent might argue against a treatment due to potential interactions, but may present this as an argument with explicit premises that physicians can challenge based on patient-specific factors. The physician might counter that the interaction risk is mitigated by the patient's genetic profile, leading the system to revise its recommendation. This will preserve clinical autonomy while providing sophisticated decision support. The AI agent serves as an argumentative partner that ensures all relevant evidence is considered, but the physician retains control over how that evidence is weighted given the specific patient context.
This is no mean feat for an agent, however, since patients' and physicians' comprehension and trust of explanations in this context have been demonstrated to be sensitive even to the \emph{format} in which this information is delivered \cite{Rago_24_Exploring}. This highlights the suitability of argumentation for this challenging task, however, since it has been shown to support explanations of different formats to adapt to users' preferences \cite{Rago_21}.

In scientific peer review, one of the problems is scale and consistency. Reviewers must evaluate whether conclusions follow from evidence, identify methodological flaws, and assess contribution significance, all while managing increasing submission volumes. This is a particularly urgent problem in AI, where the number of required reviewers does not scale linearly with the rapidly increasing submission numbers, meaning LLM assistance is already being trialed at top conferences.\footnote{\url{https://tinyurl.com/AAAI-LLM-Press-Release}} Argumentative AI agents could construct detailed argument frameworks connecting a paper's claims to its evidence, explicitly modeling the inferential steps and identifying weak links. Reviewers would see not only what the paper claims but how those claims are supposedly supported. They could challenge specific inferential steps, add missing considerations, or identify unstated assumptions. Over time, such systems could help communities converge on shared standards for what constitutes sufficient argumentative support in different subfields, and could expose systematic weaknesses (e.g.,~recurring reliance on under‑powered studies) that are hard to spot one paper at a time. Initial steps have already been made towards this goal \cite{Sukpanichnant_24}, showing its potential.

\noindent \textbf{Challenges} These applications share a pattern that defines the promise of argumentative human-AI decision-making. The AI agent's advantage lies in scale and structure, i.e., synthesizing vast amounts of information into coherent, revisable AFs and generating alternatives humans might overlook. The human's advantage lies in normative judgment and contextual sensitivity, i.e., setting goals, interpreting risks through value systems, and recognizing when formally valid conclusions are practically (or ethically) unacceptable.
The promise is to combine these complementary strengths in ways that improve decision quality while preserving human agency. Yet realizing this vision faces several challenges.

First, current benchmarks evaluate isolated tasks (e.g.,~argument extraction accuracy) but not the multi-turn interactions that will characterize the human-AI systems. Specifically, traditional evaluation metrics like accuracy scores might fail to capture whether human-AI teams make better decisions than either alone. We will need new evaluation frameworks that evaluate the following questions: (i) does the system actually save time, or does inspecting and correcting arguments take longer than working independently? (ii) do collaborative decisions show fewer errors and more robust justifications than individual human or AI decisions? (iii) how easily can humans locate and correct mistakes, and do corrections properly propagate through the reasoning chains? and (iv) does the system reduce or increase the human user's mental effort required?
\footnote{Note that while there is some work on evaluating human-AI interactions \cite{ma2024you,ma2025towards,cabitza2025too}, adapting these to contestable argumentation systems is a big open challenge.}

Preventing trust erosion is also an important challenge. AI agents must communicate the uncertainty and strength of arguments without causing over-reliance or unnecessary skepticism. For instance, humans need to know when AI-generated arguments are based on strong evidence versus speculation, when multiple models disagree, and which conclusions are most sensitive to, say, premise changes. There needs to be clearly define roles and dialogue protocols, such as who initiates the dialogue and argumentative process, how disagreements are resolved, when the dialogue should terminate, all while remaining practically feasible \cite{Modgil_17}. Agents should ideally generate arguments in a timely-efficient manner, should not consume excessive computational resources that make deployment unaffordable, and should handle sensitive data (e.g.,~legal or medical data) with privacy guarantees. Additionally, agents must be aligned with domain norms, whether it is respecting jurisdictional legal reasoning, clinical guidelines in medicine, or methodological standards in peer review. Without proper governance mechanisms, AI agents may produce arguments that appear stylistically correct but violate substantive domain requirements.

Even with these challenges, we view the trajectory outlined above as encouraging. As evidenced by recent works, we see that the technical ingredients are beginning to align. What remains is to treat argumentation not merely as a representational formalism or a set of NLP tasks, but as a design paradigm for human–AI collaboration, one in which the output is a decision, and the product is the argumentative process that justifies, qualifies, and, when necessary, changes that decision under scrutiny.

\section{Conclusion}

We have discussed how advances in argumentation framework mining and synthesis and argumentative reasoning are not isolated improvements but interconnected capabilities that enable a new paradigm: \textit{argumentative human-AI decision-making}.

The path forward requires contributions from multiple communities. AI researchers must develop architectures that support contestable reasoning and interactive revisions. HCI researchers must design interfaces that make complex argumentation accessible. Domain experts (e.g.,~scientists, doctors) must help define what constitutes good argumentative practice in their fields. Ethicists and policymakers must establish governance frameworks that ensure these systems respect human values and professional norms.

To reemphasize, the goal is not to automate human judgment but to augment it, to create AI agents that amplify human expertise rather than replace it. If successful, argumentative human-AI decision-making could transform how we approach complex problems, combining the processing power and consistency of AI agents with the 
values and contextual understanding that humans possess. The result would be decisions that are not only more accurate but more transparent, more justifiable, and more aligned with humans.



\begin{acks}
This research is partially supported by the National Science Foundation (award \#2232055)
and by the European Research Council (ERC) under the European
Union’s Horizon 2020 research and innovation programme (grant agreement \#101020934).
The views and conclusions expressed in this paper are those of the authors and do not necessarily reflect the official policies or positions of the sponsoring organizations, agencies, or governments.
\end{acks}



\bibliographystyle{ACM-Reference-Format} 
\bibliography{sample}


\end{document}